# Muscle-Inspired Magnetic Actuators that Push, Pull, Crawl, and Grasp

Muhammad Bilal Khan[1,2*], Florian Hofmann[1,2], Kilian Schäfer[1,2], Matthias Lutzi[1,2], and Oliver Gutfleisch[1,2]

[1]Functional Materials, Institute of Materials Science, Technical University of Darmstadt, 64287 Darmstadt, Germany

[2]Additive Manufacturing Center, Technical University of Darmstadt, 64289 Darmstadt, Germany

*Corresponding author: muhammad.khan3@tu-darmstadt.de

## Abstract

Functional magnetic composites capable of large deformation, load bearing, and multifunctional motion are essential for next-generation adaptive soft robots. Here, we present muscle-inspired magnetic actuators (MMA), additively manufactured from a thermoplastic/permanent magnet polyurethane/Nd$_2$Fe$_{14}$B (TPU/MQP-S) composite using laser powder bed fusion (LPBF). By tuning the laser-energy scale between 1.0 and 3.0, both mechanical stiffness and magnetic response are precisely controlled: the tensile strength increases from 0.28 to 0.99 MPa while maintaining 30-45% elongation at break. This process enables the creation of 0.5 mm-thick flexural hinges, which reversibly bend and fold under moderate magnetic fields without damage. Two actuator types are reported showing the system's versatility. The elongated actuator with self-weight of 1.57 g, magnetized in its contracted state, achieves linear contraction under a 500 mT field, lifting 50 g (32x its own weight) and sustaining performance over at least 50 cycles. Equipped with anisotropic frictional feet, it supports movement of a magnetic crawling robot that achieves up to 100% locomotion success on textured substrates. The expandable actuator exhibits reversible opening and closing under a 300 mT field, reliably grasping and releasing different objects, including soft berries and rigid 3D printed geometries. It can also anchor in a tube while holding suspended 50 g loads. This work demonstrates a LPBF-based strategy to program both stiffness and magnetization within a single material system, enabling remotely driven, reconfigurable, and fatigue-resistant soft actuators. The approach opens new possibilities for force controlled, multifunctional magnetic soft robots for adaptive gripping, locomotion, and minimally invasive manipulation of biomedical tools.

## Keywords

1. **Introduction**

Soft robotics has advanced beyond the mechanical limitations of conventional rigid-body systems by employing compliant materials that reproduce the adaptive motion and versatility of biological organisms [1-4]. Unlike traditional robots composed of discrete rigid joints and electric motors, soft robots utilize deformable architectures that serve simultaneously as structural and functional components, similar to the integrated operation of muscles, tendons, and connective tissues in living systems [1-2,5-8]. This bioinspired strategy enables safe interaction with complex or delicate environments and has found applications in biomedical devices [9-13], rehabilitation [14-15], marine exploration [16], and precision manipulation [6,13,17-18]. Pneumatic artificial muscles (PAMs) [19] exemplify this class of actuators, offering lightweight compliance and high specific power; recent pneumatic and vacuum-driven systems have introduced modular constraints and reconfigurable mechanisms capable of complex, reversible motion [20-23]. Nevertheless, reliance on external pressure sources limits miniaturization and portability. Realizing compact artificial muscles that combine large strain, tunable stiffness, and sustained force output remains a central challenge.

Among available driving principles such as dielectric [8], thermal [24], hydraulic [25-26], and pneumatic [19,22-23], magnetic actuation offers a distinctive combination of wireless control, rapid response, and deep penetration through opaque or sealed environments [27-30]. Embedding magnetic particles in elastomeric matrices yields magneto-responsive composites that deform under external magnetic fields [12,31]. Hard-magnetic elastomers with Nd-Fe-B microparticles can be pre-magnetized in desired orientations, enabling programmed deformation such as bending, twisting, or folding under uniform fields [32-34]. Earlier studies showed elastomers with spatially patterned ferromagnetic domains that undergo rapid, reversible shape morphing [11,29] and reconfigurable robots switching between crawling, rolling, and gripping purely via magnetization patterning [35-39]; magnetically driven swimmers [9,16], inchworms [33,40], and grippers [17,41] have been reported with magnetic actuation. Hybrid material systems further extend capability by coupling magnetic fillers with liquid-crystal elastomers or shape-memory polymers, enabling multi-stimuli actuation and shape locking [42-44]. Despite these advances, most magnetic soft actuators remain bending-dominated and achieve only modest contractile strains.

Achieving muscle-like axial contraction under magnetic stimulation remains a goal. Reported devices typically exhibit limited uniaxial shortening and depend on bending or torsion to generate motion [5,6,44]. Replicating the linear contraction of natural muscle fibers demands anisotropic stiffness and localized compliance that is difficult to implement in homogeneous composites. Magnetic fields also act globally, complicating selective activation of subregions within a single robot and obstruct coordinated multi-segment behaviors (e.g., simultaneous gripping and extension). Moreover, cast or laminated magnetoactive elastomers often suffer from weak interfacial bonding, prone to functional fatigue and delamination during cyclic operation. Finally, to our knowledge, conventional fabrication struggles to reproduce sub-millimeter flexural features that can be integrated into actuators geometries to get distributed compliance analogous to biological joints.

Additive manufacturing (AM) addresses these constraints by enabling complex architectures with spatially programmable properties [45-47, 57-58]. A suitable AM route is laser powder bed fusion (LPBF) of thermoplastic magnetic composites, which shows strong potential for precise spatial control of energy input and thus production of monolithic, mechanically graded magnetoactive materials in a single step [45-46,48]. Recent work on TPU/Nd-Fe-B (TPU/MQP-S) composites showed that varying the laser-energy scale locally tunes stiffness without significantly affecting elasticity, producing graded mechanical and magnetic profiles and improved composite packing [45-47]. Importantly, this materials system leverages $Nd_2Fe_{14}B$-type permanent magnets, which exhibit the highest energy density among commercially available magnets at and slightly above room temperature and are therefore the material of choice for high-performance motors, generators, and compact mobile devices [56]. While LPBF is shown to produce complex geometries, its potential to fabricate actuators capable of doing load bearing work has not been fully investigated. More specifically, a single, additively manufactured platform uniting linear contraction, directional locomotion, and adaptive gripping within one continuous material body has not been achieved.

This paper leverages the convergence of magnetoactive composites, sub-millimeter hinge geometry for folding mechanisms in 3D printed actuators, and LPBF additive manufacturing to produce a single-material platform capable of multiple actuation modes. We print thermoplastic polyurethane (TPU) mixed with spherical and magnetically isotropic $Nd_2Fe_{14}B$ microparticles via LPBF and use the laser-energy scale (1.0-3.0) as a design variable to tune stiffness and magnetic response. Two architectures are realized from the same composite: (i) an elongated, muscle-inspired actuator for axial contraction and load lifting, and (ii) a radially expandable actuator for reversible opening/closing to grasp and anchor. Embedding 0.5 mm-thick flexure hinges and programming magnetization profiles enables contraction, crawling (with anisotropic-friction feet), and gripping/anchoring within a compact, monolithic system. Quantitatively, the process-property-performance linkage is explicit: tensile strength increases from 0.28 to 0.99 MPa across the printing energy range while preserving 30-45% strain at break; the elongated actuator can repeatedly lift 50 g (approx. 23-32× self-weight) and maintains performance over 50 cycles; and when employed in a worm-like crawler, the crawler achieves 90% crawling success on SiC paper and 100% on a tissue substrate. This study advances magnetic artificial muscles along three axes: 1. Process-encoded mechanics and magnetics: LPBF energy scaling as a continuous design parameter to tune stiffness, porosity, and magnetic output; 2. Miniaturized compliant architecture: realization of sub-mm flexural hinges within a magnetoactive printed composite for high-cycle, load-bearing actuation; 3. Multimodal function: integration of anisotropic friction feet and programmed magnetization to achieve crawl and grip operation across surfaces and object geometries, without multi-material assembly or complex reconfiguration steps. Together, these elements establish a material-centric pathway to adaptive, untethered soft machines for locomotion, manipulation, and anchoring in real-world environments.

## 2. Materials and Methods

### 2.1 Bioinspired Design Concept

The actuator systems developed in this work were conceived following the biomechanical principles of natural muscle actuation, emphasizing the contraction-relaxation behavior of human skeletal muscles and the distributed deformation of human fingers during grasping (Figure 1A). Inspired by these biological mechanisms, two functionally distinct but compositionally identical actuator types were designed: (i) an elongated actuator, which produces uniaxial contraction resembling a linear muscle fiber, and (ii) an expandable actuator, which generates radial opening and closing motions for grasping-like behavior (Figure 1B). Both designs employ the same magnetoactive material and laser powder bed fusion (LPBF) process, enabling multi-modal magnetic actuation within a single material platform. The optical images and functional demonstrations shown in Figure 1C-D summarize the concepts and experiments presented throughout this study (see Supplementary movie M1).

## 2.2 Materials and Composite Preparation

The magnetoactive composite was formulated using thermoplastic polyurethane (TPU; Flexa Grey, Sinterit, Poland; 20-105 µm) as a flexible binder and spherical $Nd_2Fe_{14}B$ microparticles (atomized MQP-S, Neo Magnequench, Singapore; 35-55 µm) as a hard-magnetic filler (Figure 2A). Both powders were mixed for one hour in a Turbula WAB mechanical mixer (Switzerland) to ensure homogeneous dispersion of magnetic particles within the TPU matrix. The resulting mixture contained 50 wt % magnetic filler, corresponding to ≈12.7 vol %. The smooth surface morphology and spherical particle shape of the filler promoted high flowability and integrated with binder, improving layer uniformity during LPBF fabrication.

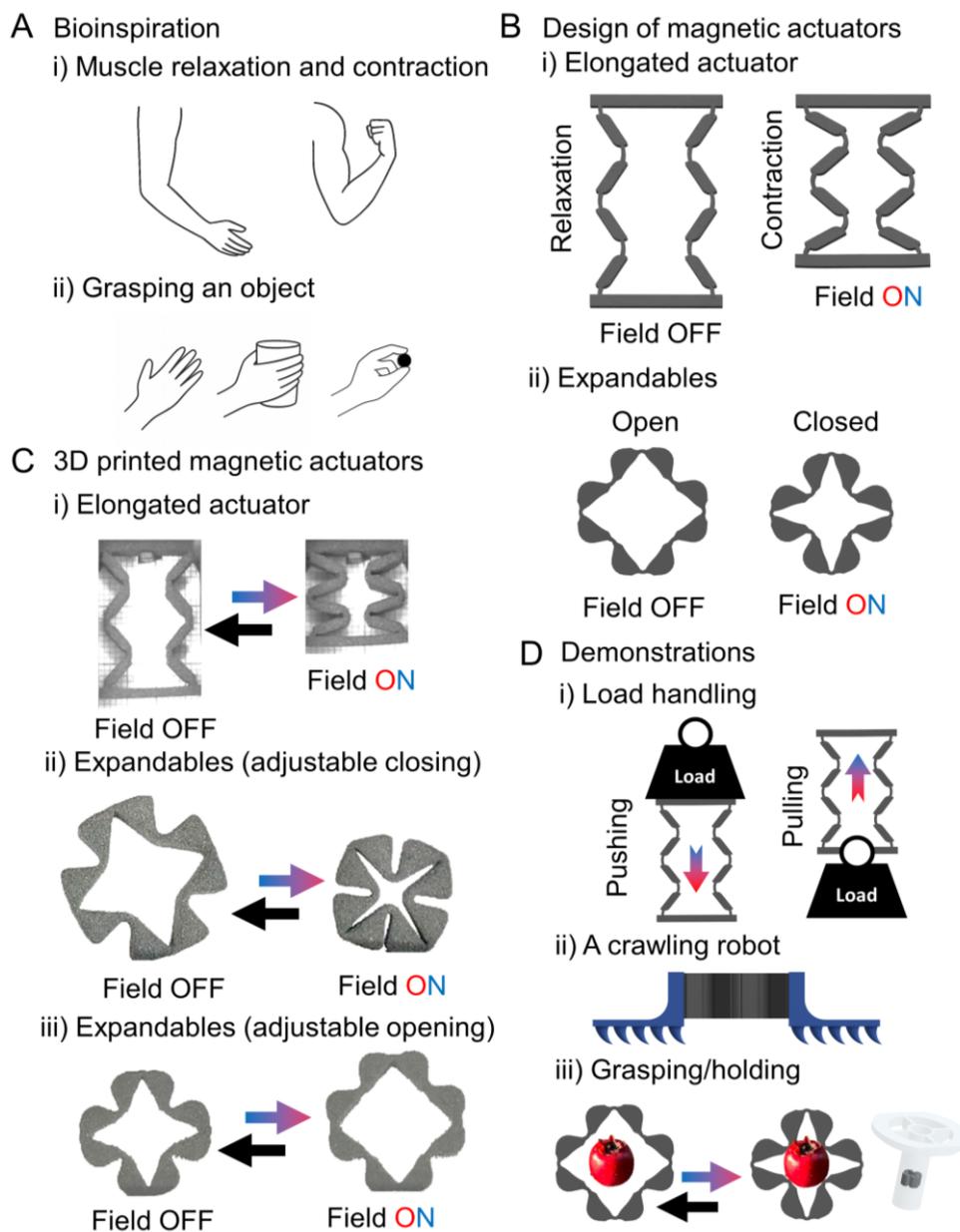

**Figure 1. Bioinspired design and actuation concept of 3D printed magnetic actuators. A)** Bioinspired concept. *(i)* Schematic illustration of a human arm showing the biceps muscle in its relaxed and contracted states, indicating the natural contraction-relaxation mechanism of muscle tissue. *(ii)* Simplified schematic of a human hand illustrating grasping and holding of objects, emphasizing the principle of distributed muscle actuation for coordinated deformation. **B)** Design principle of muscle-inspired magnetic actuators with flexure hinges. *(i)* An elongated actuator capable of reversible contraction under an external magnetic field, mimicking linear muscle actuation. *(ii)* Expandable magnetic actuators programmed to exhibit reversible opening and closing motions upon magnetic stimulation. The folding direction and magnitude of each actuator are determined by the programmed magnetization profiles. **C)** Optical images of 3D printed magnetic actuators in various actuation states. *(i)* Elongated actuator demonstrating magnetic-field-induced contraction under 500 mT. *(ii, iii)* Expandable actuators showing magnetically controlled closing and opening behavior under 300 mT, each designed with distinct printing formats and magnetization directions for tailored deformation. **D)** Overview of demonstrations presented in this study. *(i)* The elongated actuator performs linear actuation to push and pull external loads, and its mechanical performance is quantitatively characterized. *(ii)* The same actuator is integrated as the artificial muscle in a soft crawling robot equipped with non-magnetic anisotropic frictional feet, enabling directional locomotion across different surfaces under remote magnetic control. *(iii)* Expandable actuators are applied for object manipulation, grasping objects with diverse surface textures and geometries, and forming an internal grip within a confined pipe while supporting external loads under magnetic actuation.

## 2.3 Additive Manufacturing via Laser Powder Bed Fusion

All actuators were printed using a laser powder bed fusion (LPBF) system (Sinterit Lisa Pro, Poland). As illustrated in Figure 2B, the process employs a mirror-guided laser beam to selectively melt and solidify successive powder layers, gradually building up the actuator geometry directly from the composite feedstock. Key LPBF parameters were kept constant except for the laser-energy scale, a dimensionless factor varied between 1.0 and 3.0 to tune the mechanical stiffness and magnetic remanence of the printed specimen. Higher laser energy promoted interlayer densification and reduced porosity, leading to mechanically stronger yet flexible composites. Tensile stress increased from 0.28 MPa (1.0) to 0.99 MPa (3.0), confirming controllable stiffness adjustment through printing parameters (Figure 2C). A representative elongated actuator fabricated under optimized conditions (LP 2.6) is shown in Figure 2D.

## 2.4 Actuator Design and Geometry

### 2.4.1 Elongated Actuator

The elongated actuator was designed to mimic the axial contraction of muscle fibers through a zig-zag hinge structure linking two rigid end plates (Figure 1B i). Each actuator incorporated 0.5 mm-thick flexural hinges that bend under magnetic torque while maintaining structural continuity. The total height (27.5 mm), hinge spacing, and angle ($\beta = 113°$) were adjusted via computer aided design (CAD) modelling to balance compliance and strength. Detailed labeled schematics and full dimensions are provided in *Supplementary Figure S1* and *Table S1*. These sub-millimeter hinges for soft actuators represent one of the smallest 3D printed hinge scales achieved in magnetically responsive structures using powder bed fusion, enabling large, reversible deformation without mechanical degradation under defined contraints.

### 2.4.2 Expandable Actuators

Two expandable actuator variants were designed for reversible radial motion: an opening configuration and a closing configuration. Both contain a four-lobed, radially symmetric architecture with inner and outer hinge sets (0.5 mm thick, 10 mm tall) arranged around a central cavity (Figure 1B ii). The folding behavior is dictated solely by the magnetization orientation, one variant expands outward upon field application, the other contracts inward, allowing bidirectional control within identical geometries. Complete geometric data, including hinge lengths, inner and outer angles, and diameters (18.5-22.5 mm), are summarized in *Supplementary Figure S2* and *Table S2*.

## 2.5 Magnetization of Actuators

Actuators were magnetized using a pulsed field magnetometer with a field strength of 3 T. The elongated actuator was magnetized in its fully contracted configuration to encode a field-responsive profile inducing contraction under a uniform external field (Figure 3A i). The expandable actuators were magnetized in mechanical jigs defining their rest geometries (open or closed) to determine the folding direction shown in Figure 5A. Precise alignment of the magnetization axes was ensured by non-magnetic fixtures that constrained deformation during pulsed-field exposure.

## 2.6 Magnetic Actuation Testing

### 2.6.1 Elongated Actuator Performance

Magnetic actuation was characterized using a custom 3D printed holder mounted between the poles of a water-cooled electromagnet (Figure 3A ii). The Prusa SL1s resin printer was used for the manufacturing of the actuator holder as well as to 3D print frictional feet using the hard-tough black resin - H200, from eSun. For pulling tests, the actuator's upper base was fixed in the holder cavity and loads of 20-50 g were suspended from the lower base. For pushing tests, the actuator base was fixed and loads were placed on its top plate. Actuation fields up to 500 mT were applied, and displacement was recorded by a digital camera. Cyclic actuation tests under a 100 g load were performed for 50 cycles to assess durability (Figure 3C ii). For locomotion experiments, the elongated actuator was integrated into a magnetic crawling robot equipped with non-magnetic 3D printed feet containing asymmetric pointy surface micro-textures that induce anisotropic friction [50]. Alternating contraction and relaxation under periodic magnetic-field switching generated directional crawling across different substrates (Figure 4 i-iii). Crawling efficiency was quantified by measuring forward displacement and success rate across SiC abrasive paper, latex, and laboratory tissue (Figure 4 iv).

### 2.6.2 Expandable Actuator Performance

Expandable actuator experiments were conducted in the same electromagnet setup (Figure 5B). A magnetic field of 300 mT was applied while deformation was recorded optically. The actuators were evaluated for object handling, grasping, and anchoring tasks (Figure 5C-D). Thirteen representative objects, including wild berries, geometrically complex items, and 3D printed tubes, were grasped, lifted, and released in triplicate trials (Figure 5C and Figure 6A). For anchoring tests, a Halbach-array magnet was used to generate a localized radial field within a UV resin-based 3D printed tube. Upon activation, the actuator formed a conformal grip on the inner wall and supported suspended loads up to 50 g, demonstrating stable magnetic anchoring (Figure 6 B). Detailed structural and dimensional information for both expandable designs is provided in *Supplementary Figure S2* and *Table S2*.

### 2.7 Mechanical Testing

Uniaxial tensile tests were performed on printed dumbbell shaped samples (ASTM D638) using a an Inspekt Tab. 5 (Hegewald and Peschke, Germany) testing machine at a speed of 5 mm min$^{-1}$. Engineering stress-strain curves were used to derive tensile stress values, confirming an increase from 0.28 MPa at laser-energy scale 1.0 to 0.99 MPa at 3.0 (Figure 2C). These measurements were to verify that LPBF processing directly governs both stiffness and actuation strength by controlling local densification and particle bonding as it has been established in previous studies [45-46].

## 2.8 Microscopy and Imaging

Scanning electron microscopy (SEM; Tescan VEGA3-SBH) was employed to examine TPU and Nd-Fe-B morphologies before and after mixing (Figure 2A). Optical digital imaging was used to record actuator deformation, hinge motion, and fatigue behavior.

## 3 Results and Discussion

The magnetoelastic actuators developed here demonstrate how additive manufacturing parameters can be used to affect magnetic responsiveness and mechanical strength in order to realize distinct motion modes such as contraction, crawling, and gripping, within a single material system. The integration of programmable magnetization with tunable mechanical stiffness enables multifunctional soft actuation driven entirely by remote magnetic fields.

## 3.1 Production and Nature of the Magnetic Composites

The magnetoactive composite was prepared from thermoplastic polyurethane (particle size 20-105 μm) as a flexible binder and spherical $Nd_2Fe_{14}B$ magnetic powder (particle size 35-55 μm) as a hard-magnetic filler. SEM imaging showed the uniformly spherical magnetic particles, ensuring excellent flowability and dispersion across the powder bed. Fabrication was conducted using laser powder bed fusion (LPBF), where successive layers of polymer were selectively melted and solidified by a scanning laser beam. The laser-energy scale, a dimensionless parameter defining the relative laser power, was varied between 1.0 and 3.0, while all other parameters were fixed (printing parameters are available is *Supplementary Table S3*). Increasing the energy scale promoted interlayer fusion and reduced porosity, producing denser, mechanically stronger composites while maintaining flexibility. Mechanical testing revealed a near-linear increase in tensile strength from 0.28 MPa (energy scale 1.0) to 0.99 MPa (3.0). The strain at break remained between 30-45 %, confirming that elasticity was preserved despite increased stiffness. These results are consistent with previously reported densification behavior for the same TPU/MQP-S system [45], where density increased from 1.1 to 1.4 g cm$^{-3}$ and porosity decreased from 33% to 14%, accompanied by a remanence rise from 47 mT to 76 mT and a nearly constant coercivity of ≈ 885 mT. Together, these trends show that LPBF processing parameters directly couple the mechanical and magnetic performance of the printed composite.

Each actuator design incorporated 0.5 mm-thick flexural hinges, a sub-millimeter structural scale comparable to morphing elements reported in butterfly-inspired systems [47]. These thin hinges allow reversible bending and folding under magnetic torque without mechanical failure, representing uniquely small printed hinge thicknesses achieved in magnetic actuator systems. The actuator printed at an energy scale of 2.6 provided the most balanced combination of stiffness and compliance, showing smooth hinge transitions and no interlayer delamination.

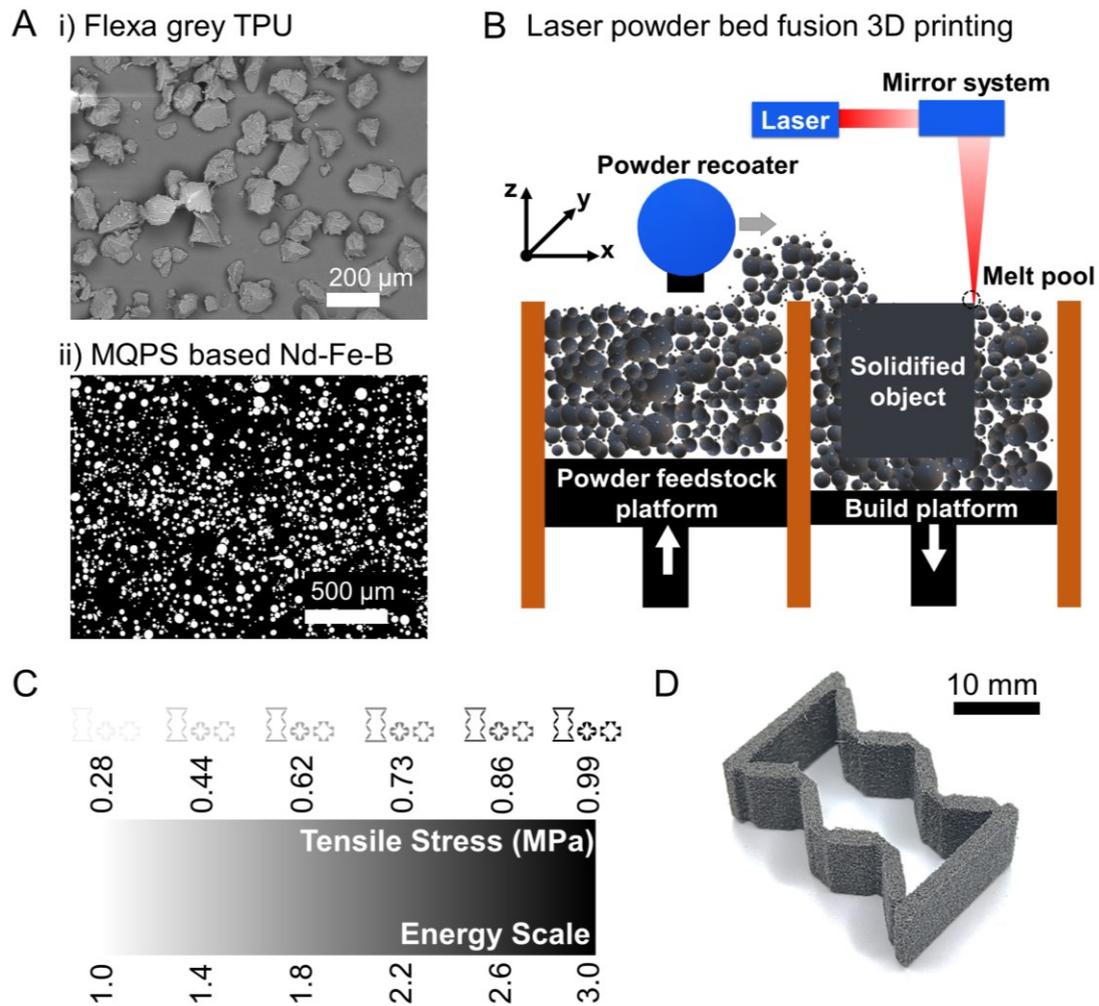

**Figure 2. Material morphology, fabrication, and tunability of 3D printed magnetic actuators. A)** Microstructural characterization of constituent materials. *(i)* Scanning electron microscopy (SEM) image of the flexa grey thermoplastic polyurethane (TPU) matrix used as the elastomeric binder. *(ii)* SEM image of spherical $Nd_2Fe_{14}B$ magnetic microparticles used as the magnetic filler. **B)** Schematic illustration of the laser powder bed fusion (LPBF) 3D printing process employed for actuator fabrication (overview inspired from [49]). A laser beam, guided by a mirror system, selectively melts and solidifies the magnetoactive composite powder in a layer-by-layer manner. Here, key components include the powder feedstock platform, recoater

blade, melt pool, solidified layer, and build platform, showing the additive buildup of a shape directly from magnetic composite feedstock. **C)** Tailoring of mechanical and magnetic properties by tuning laser energy input. The tensile stress of 3D printed magnetic actuators increases with higher laser energy density (1.0 - 3.0), demonstrating process-dependent control of resulting material performance. **D)** Optical photograph of a representative elongated 3D printed magnetic actuator fabricated from the optimized magnetoactive composite, showing preserved design details e.g. the 0.5 mm hinges.

### 3.2 Magnetically driven contraction and load-bearing performance

The elongated actuator was magnetized in its fully contracted configuration under a 3 T pulsed field to encode a magnetization profile that drives axial contraction when exposed to a uniform magnetic field (Figure 3A i). Actuation tests were performed by placing the sample vertically between the pole shoes of a water-cooled electromagnet using a custom 3D printed holder (Figure 3A ii). Loads between 20 and 50 g were applied to quantify both pulling and pushing modes. In pulling tests, the displacement under load decreased gradually with increasing laser energy scale. Under a 20 g load, actuators printed at LP 1.0 and LP 3.0 exhibited contractions of approximately 10 mm and 5 mm, respectively. Increasing the load to 50 g reduced displacement slightly ($\approx$ 1-2 mm) across all samples, indicating consistent mechanical stability. In pushing tests, stiffer actuators exhibited improved recovery: LP 1.0 recovered incompletely, while LP 3.0 recovered $\approx$ 80 % of its original length. The actuator produced at LP 2.6 achieved the optimal compromise, displaying $\approx$ 7 mm contraction with $\approx$ 86 % recovery, yielding the highest effective displacement (Figure 3C i).

The self-weight of each actuator and its corresponding lift ratio for a 50 g load were:

- LP 1.0: 1.57 g ($\approx$ 31.8× self-weight)
- LP 1.4: 1.72 g ($\approx$ 29.1×)
- LP 1.8: 1.86 g ($\approx$ 26.9×)
- LP 2.2: 2.01 g ($\approx$ 24.9×)
- LP 2.6: 2.17 g ($\approx$ 23.0×)
- LP 3.0: 1.96 g ($\approx$ 25.5×)

Even the densest actuator lifted more than 23 times its own weight, while the most compliant variant (LP 1.0) lifted over 30 times its weight. Here, lifting not only refers to holding the weight but also the repeatable contraction-elongation cycles. Despite repeated bending at 0.5 mm hinges under load, no cracking or delamination was observed. Under cyclic loading of 100 g for 50 actuation cycles, the actuators retained over 60% of their displacement amplitude, demonstrating exceptional fatigue resistance (Figure 3C ii). The success rate for repeatable actuation rose from 60% at LP 1.0 to 100% for LP 1.4-3.0, confirming that improved layer fusion and reduced porosity directly enhance fatigue durability.

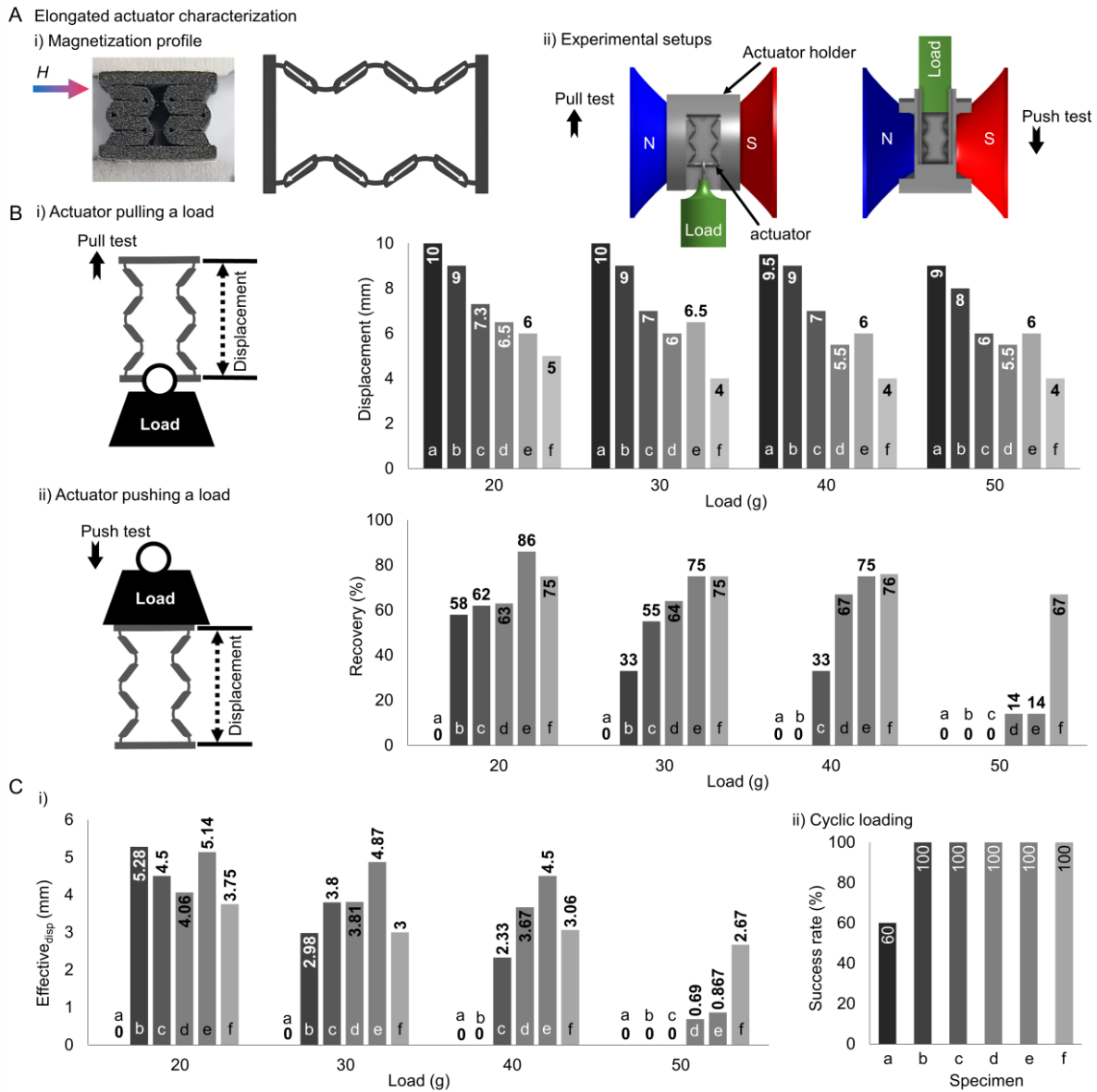

**Figure 3. Characterization and performance of elongated muscle-inspired magnetic actuators. A)** Characterization and experimental setup. *(i)* Magnetization profile of the elongated actuator magnetized in its fully contracted state, enabling contraction upon application of a uniform external magnetic field. *(ii)* Experimental setups for pulling and pushing tests. A custom-built 3D printed actuator holder was mounted between the pole shoes of an electromagnet. For pulling tests, the actuator's top base was fixed in the holder cavity and a load was suspended from the lower base. For pushing tests, the actuator's base plate was fixed while the load was mounted on its top end. Depending on the programmed magnetization direction, the actuator either pulled or pushed the load under magnetic stimulation. **B)** Actuation performance under magnetic loading. *(i)* Displacement of actuators during pulling tests at different applied loads (20-50 g). Samples a-f correspond to different laser energy scales (LP 1.0-3.0) used during 3D printing. Higher laser energy yields stronger actuation, as shown by larger displacements. *(ii)* Pushing test results presented as recovery percentage for the same load range, confirming field-dependent, reversible deformation of the printed magnetic actuators. **C)** Mechanical durability and actuation consistency. *(i)* Effective displacement of elongated actuators summarizing combined pulling and recovery behavior. *(ii)* Cyclic loading test under a 100 g load for 50 cycles demonstrating stable magnetic actuation without loss of performance or structural degradation, confirming robustness of the 3D printed magnetoactive material.

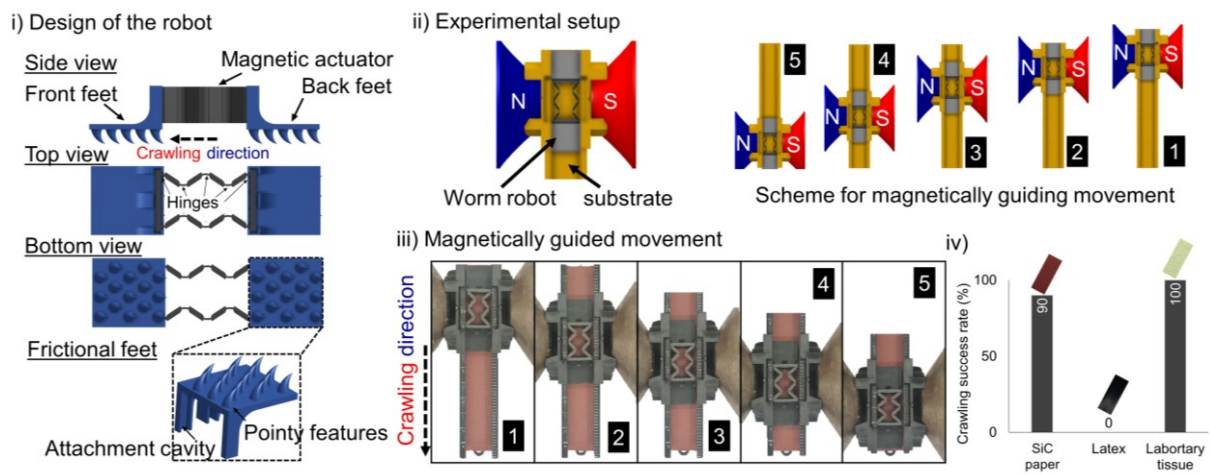

**Figure 4. Magnetic crawling robot driven by an elongated actuator.** *(i)* Design schematic showing the actuator integrated with two non-magnetic 3D printed foot segments patterned with asymmetric surface features to provide directional (anisotropic) friction, allowing movement in one direction while anchoring in the opposite. *(ii)* Experimental setup in which the robot, placed on a substrate mounted on a guiding rail inside an electromagnet, is actuated by periodic magnetic-field switching to achieve alternating contraction and extension. *(iii)* Time-lapse images demonstrating controlled, directional locomotion. *(iv)* Crawling success rate on substrates of different surface textures (SiC paper, latex, and laboratory tissue), showing the role of substrate frictional properties in locomotion efficiency.

## 3.3 Magnetically controlled crawling locomotion

When combined with two non-magnetic foot segments containing asymmetric micro-features, the elongated actuator served as the muscle element of a magnetic crawler. These feet introduced anisotropic friction [50-52], anchoring in one direction and sliding in the opposite, thus converting the actuator's reciprocal deformation into net forward movement (Figure 4 i-iii). The movement is similar to the inching of an inchworm that uses its feet to adhere to the crawling surface while using the body elongation-contraction cycles for directional movement [53-54].

Crawling performance of the magnetic crawler strongly depended on the substrate texture (Figure 4 iv). Each surface was tested three times, and the crawling success rate was defined as the percentage of successful, complete unidirectional crawls. On SiC abrasive paper, the crawler achieved 90% success, completing three full cycles and advancing ≈ 60 mm in total displacement without slippage. On latex, the lack of directional friction led to failure, while on laboratory tissue, the fibrous texture enabled consistent 100% success. These results demonstrate that the combination of surface friction, construct of the feet features, and actuator compliance determines locomotion efficiency, and that tailoring foot microgeometry can enhance traction on various surfaces.

## 3.4 Expandable actuators for gripping and anchoring

The expandable actuator converted magnetic torque into radial folding motion, producing reversible opening and closing analogous to biological grasping (Figure 5A) [55]. Without a magnetic field, the actuator remained open; upon field application, it symmetrically closed. The actuation was observed between the same electromagnet poles, and deformation was recorded by a digital camera (Figure 5B).

The actuator successfully grasped, lifted, and transported 13 distinct objects (Figure 5C i), including six types of wild berries and seven geometrically diverse items (3D printed letters, pyramids, domes, cubes, chalk, a straw, and a pipe-like structure) with a 100% success rate. The actuator's inner cavity accommodated objects of varying diameters, confirming its ability to conform to various geometries (Figure 6). The typical operation involved fixing an object within the cavity, transporting it magnetically, and releasing it upon field-induced reopening (Figure 6A ii).

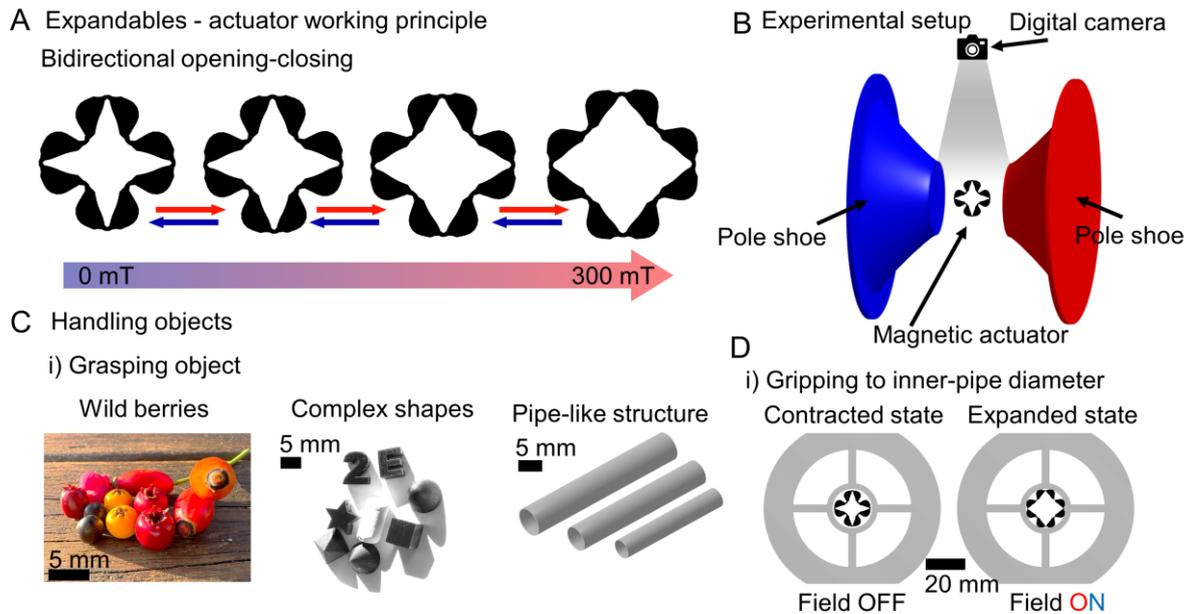

**Figure 5. Working principle and experimental setup with test scenarios for expandable magnetic actuators. A)** Working principle of the expandable magnetic actuator. The actuator exhibits reversible opening and closing in response to external magnetic fields, enabling a muscle-like grasping motion. In the absence of a magnetic field, the actuator remains open; under magnetic stimulation, the programmed magnetization pattern induces symmetric folding for closing. **B)** Experimental setup for magnetic actuation testing. The actuator is positioned between the pole shoes of an electromagnet, and a digital camera records its field-induced deformation in real time for displacement and angle analysis. **C)** Object handling of different types. *(i)* To be grasped diverse objects including wild berries, irregularly shaped items with complex contours or sharp edges, and cylindrical 3D printed tubes of various diameters. **D)** Magnetic anchoring within tubular geometries. *(i)* Schematic illustration of the actuator forming a radial grip against the inner wall of a pipe upon magnetic activation.

Beyond external grasping, the same actuator design enabled anchoring inside tubular geometries (Figure 6B). When exposed to a magnetic field, the actuator expanded radially to grip the inner wall and released upon deactivation. In tests using a Halbach-array magnet, the actuator supported a 50 g suspended load without detachment (Figure 6B iii). Such remotely controlled anchoring capability suggests potential (after scaling down/up actuator geometry) for biomedical or industrial applications where soft, adaptive fixation is required in confined environments.

While the actuators demonstrated reliable and repeatable performance, several current limitations point toward promising directions for future refinement. Presently, actuation is open-loop, and deformation is predefined by the actuator's geometry and magnetization profile. Object grasping is tuned to expected dimensions rather than sensed contact. Incorporating embedded strain or pressure sensors (e.g., printed resistive or capacitive networks) would enable closed-loop force control. Measuring real-time grasping forces and object compliance would allow adaptive gripping of deformable or delicate targets. It is also to note that the lower laser-energy scales yield high strain but limited recovery; higher scales enhance recovery but reduce deformation amplitude. In the future, implementation of spatially graded LPBF, locally varying energy input to co-print compliant hinge regions with stiff structural zones can be considered. Combining this with controlled magnetization gradients can further enhance actuation efficiency and structural resilience.

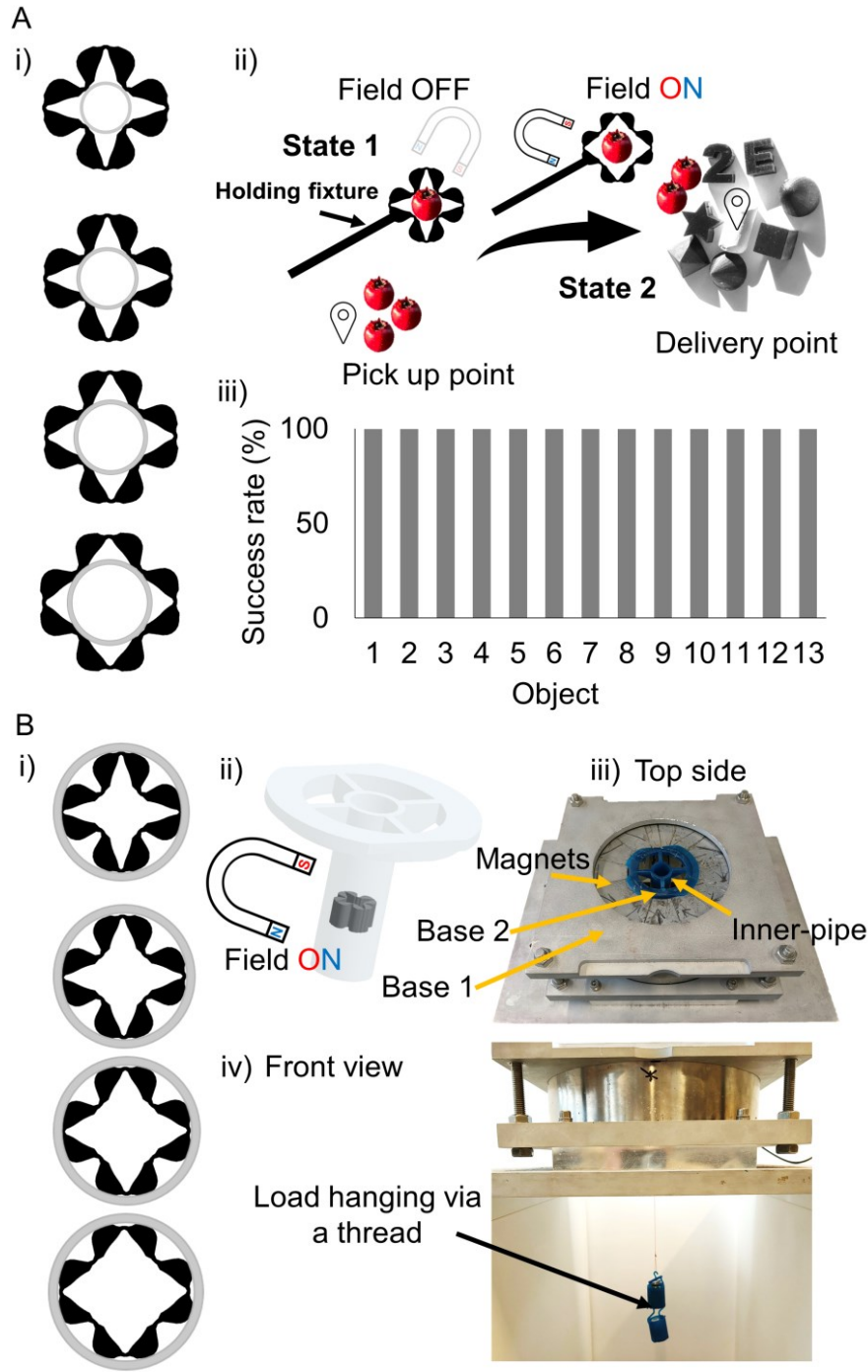

**Figure 6. Expandable magnetic actuators for object manipulation and anchoring applications. A)** Object handling of different types. *(i)* Schematic representation of the internal working space, showing the adaptable cavity capable of accommodating objects of varying size and geometry. *(ii)* Illustration of the object manipulation workflow: the actuator first secures an object within its inner cavity, transfers it to another location, and releases it upon magnetic-field activation, which reopens the actuator. *(iii)* Measured success rate for grasping, lifting, and releasing 13 distinct objects without slippage or drop, confirming reliable and repeatable

operation. **B)** Magnetic anchoring within tubular geometries. *(i)* Diagram of the external working range, showing the actuator's ability to conform to pipes of different diameters by adjusting magnetic-field strength. *(ii)* Schematic illustrating how remote magnetic-field control allows axial translation of the actuator inside the tube while maintaining secure anchoring. *(iii)* Photograph of the experimental setup using a Halbach-array magnet surrounding a tube containing the actuator, where the actuator supports a suspended 50 g load without losing adhesion, demonstrating its potential for remote anchoring and biomedical soft robotic applications.

## 4 Conclusion and Outlook

This study establishes a comprehensive route to muscle-inspired magnetic actuators fabricated by laser powder bed fusion of a TPU/Nd-Fe-B composite. By varying the laser-energy scale between 1.0 and 3.0, both mechanical and magnetic properties were tuned, yielding programmable deformation under moderate magnetic fields. The elongated actuator demonstrated reversible contraction and recovery while lifting loads up to 50 g, equivalent to 23-32× its self-weight, and maintained repeatability over 50 actuation cycles without structural loss. When coupled with anisotropic frictional feet, it achieved controlled crawling across different substrates with surface-dependent efficiency. The expandable actuator showed multifunctional capabilities such as grasping, transporting, and anchoring, handling 13 diverse objects and supporting suspended loads of 50 g inside tubular environments. A central novelty of this work lies in the integration of sub-millimeter flexural hinges, tunable stiffness, and programmable magnetization within a single, additively manufactured material. These actuators require no onboard power or wiring, operating solely under remote magnetic fields, offering a lightweight and fully enclosed design ideal for compact or sealed robotic systems. However, current actuation remains position-based, with deformation preset for known geometries. For advanced robotic manipulation, force control will be essential. Future efforts will therefore focus on: (i) embedding soft sensing elements to monitor strain or pressure during actuation, (ii) performing force-displacement calibration under applied magnetic fields, and (iii) implementing closed-loop controllers (force, impedance) to achieve dynamic, adaptive grasping of soft or irregular objects. Further research will explore multimaterial or locally graded LPBF to co-print compliant and rigid segments, integrate co-designed magnetization-stiffness maps, and miniaturize actuators for untethered operation with compact permanent-magnet sources. The possibility to integrate biocompatible TPUs also opens paths toward biomedical applications, such as minimally invasive anchoring, soft tissue manipulation, and deployable robotic instruments for confined or fluidic environments. More broadly, this work contributes towards developing adaptive, multifunctional soft machines capable of locomotion, grasping, and anchoring, governed entirely by remote magnetic control.

## Acknowledgement

This work was financially supported by the Deutsche Forschungsgemeinschaft (DFG, German Research Foundation), Project ID No. 405553726, TRR 270 and the RTG 2761 LokoAssist (Grant no. 450821862).

## Data Availability Statement

The supplementary video is available at https://zenodo.org/records/19329955, where as additional data available on request from the authors.